\documentclass[11pt,a4paper]{article}
\usepackage[preprint]{acl}
\usepackage{times}
\usepackage{latexsym}
\usepackage{amsmath}
\usepackage{amssymb}
\usepackage{dsfont}
\usepackage{times}
\usepackage{latexsym}
\usepackage[T1]{fontenc}
\usepackage[utf8]{inputenc}
\usepackage{microtype}
\usepackage{inconsolata}
\usepackage{graphicx}
\usepackage{algorithm}
\usepackage{algpseudocode}
\usepackage{booktabs}

\usepackage{microtype}

\title{Deep Language Geometry:\\Constructing a Metric Space from LLM Weights}

\author{Maksym Shamrai \\
  Institute of Mathematics of NASU \\
  MacPaw \\
  Kyiv, Ukraine \\
  \texttt{mshamrai@macpaw.com} \\\And
  Vladyslav Hamolia \\
  MacPaw \\
  Kyiv, Ukraine \\
  \texttt{vhamolya@macpaw.com} \\}

\date{}

\begin{document}
\maketitle
\begin{abstract}
We introduce a novel framework that utilizes the internal weight activations of modern Large Language Models (LLMs) to construct a metric space of languages. Unlike traditional approaches based on hand-crafted linguistic features, our method automatically derives high-dimensional vector representations by computing weight importance scores via an adapted pruning algorithm. Our approach captures intrinsic language characteristics that reflect linguistic phenomena. We validate our approach across diverse datasets and multilingual LLMs, covering 106 languages. The results align well with established linguistic families while also revealing unexpected inter-language connections that may indicate historical contact or language evolution. The source code, computed language latent vectors, and visualization tool are made publicly available at \url{https://github.com/mshamrai/deep-language-geometry}.
\end{abstract}

\section{Introduction}

Languages are complex systems with rich internal structures and dynamic evolution. Traditional linguistic classifications based on typological features, historical migration patterns, or lexical similarity have long served to group languages into families such as Indo-European, Uralic, and Turkic. However, these approaches typically capture only historical or static aspects of language similarity, potentially overlooking modern linguistic influences driven by technology and globalization. In an era where language use and structure are continuously reshaped, it is timely to develop methods that automatically capture both historical and current linguistic characteristics.

Recent advances in Natural Language Processing (NLP) have been largely driven by Large Language Models (LLMs), which have demonstrated remarkable capabilities in language modeling and a wide range of linguistic tasks \cite{devlin2018bert, radford2019language}. These models, trained on vast multilingual corpora, learn representations that implicitly encode a wide variety of lexical, syntactic, and even phonological properties \cite{conneau2019unsupervised}. 

Building on prior work \cite{shamrai2024language}, which empirically shows that the internal activations of LLM weights vary with the language of the input data, we hypothesize that the internal weights of LLMs encode valuable information about inter-language similarity and can serve as a foundation for quantifying relationships between languages.

Therefore, in this work, we propose a novel approach for constructing a metric space of languages by leveraging the weights of modern LLMs. Our method extracts high-dimensional vector representations from LLM weights activations, where the distance between any two vectors reflects the similarity between the underlying linguistic structures. Activations encode patterns of co-occurrence and contextual relationships specific to each language's grammar and lexical properties. 

We construct a metric space $(X, d_h)$, where $X$ is the set of high-dimensional language vectors and $d_h$ is the Hamming distance between them. We then design a distance-preserving mapping that projects these high-dimensional vectors into a low-dimensional space $(Y, d_{e})$, where distances are induced from the Euclidean (L2) norm. 
This transformation provides deeper insight into the latent structures encoded by LLMs.

Furthermore, we calculate this latent representation for 106 languages. This revealed the opportunity to visualize, cluster and analyze the relationships between the languages.  

Our code, computed language latent vectors, and analysis tool are made publicly available, designed to assist researchers and practitioners in linguistic analysis and offering valuable resource for further linguistic investigation.

The contributions of this work are as follows:
\begin{itemize}
    \item We introduce a novel approach that constructs a metric space of languages using LLM weights and apply it to 106 languages, enabling automatic and data-driven measurement of linguistic distances.
    \item We demonstrate that the derived metric space supports meaningful clustering of languages, reflecting both historical relationships and modern linguistic features.
    \item We fully open-source our work along with a tool for preliminary analysis. 
\end{itemize}

While not claiming linguistic expertise, this study introduces a novel toolset intended to support linguistic research.  
It offers a fresh view of language similarity by exploiting the latent knowledge embedded in LLMs.

\section{Related Work}

The quantification of language similarity has a rich history, beginning with early lexical approaches. Pioneering work \cite{swadesh1952lexico} established methods for comparing languages using shared cognates, a practice later refined by \citet{holman2011automated}, which employs normalized Levenshtein distances over fixed word lists. Although these lexical methods have been successfully used to construct language family trees, they are handcrafted and require manual effort to select and curate appropriate word lists and features.

Also, resources such as the World Atlas of Language Structures \cite{wals2005} offer comprehensive typological data that allow languages to be represented as feature vectors. Distance measures computed over these vectors have been shown to reveal groupings consistent with established genetic relationships \cite{o2016syntactic, de2024exploring}. However, these methods are limited by the quality and coverage of available databases, their reliance on expert-curated features, and their inability to fully capture language-specific variations or recent evolutionary trends.

Phonological properties offer another valuable dimension for language comparison. Studies utilizing phoneme inventory data from resources like PHOIBLE \cite{moran2014phonological} demonstrate that phonological distances -- often measured by overlap indices such as the Jaccard similarity -- can capture both genetic relationships and areal phenomena. But phonological methods need reliable phoneme lists, are affected by how sounds are written, and often miss language structure beyond sounds.

Recent deep‑learning work has popularised embedding‑based measures of language distance. Multilingual encoders such as mBERT \cite{devlin2018bert}, XLM‑R \cite{conneau-etal-2020-unsupervised} and LASER \cite{artetxe-schwenk-2019-massively, heffernan-etal-2022-bitext} produce contextual token embeddings that implicitly encode lexical, syntactic and semantic features. LASER is trained to output a single sentence vector directly, whereas mBERT and XLM‑R require a pooling step (e.g., mean pooling or the \texttt{[CLS]} token) to obtain a sentence‑level embedding. When sentence embeddings are averaged over large, balanced corpora, the resulting language‑level representations have proved useful for quantifying cross‑lingual similarity \cite{rama2020probing}. However, because the underlying encoders operate at the token -- and therefore sentence -- level, their effectiveness still depends on corpus size and domain balance.



Overall, the literature on language distance metrics has evolved from classical lexicostatistical methods and handcrafted feature extraction to sophisticated neural representations. Each approach offers valuable insights into the relationships between languages, but they often suffer from labor-intensive preprocessing, limited database coverage, or sensitivity to input variations. This motivates our approach: rather than relying on manually curated features or sentence-based embeddings, we propose an automatic, data-driven method that leverages the internal weights of modern LLMs to construct a metric space of languages. 

Moreover, to best of our knowledge, no study has attempted to derive a language metric space from decoder‑only LLMs. The method introduced here is therefore the first to use weight‑level signals in causal transformers for measuring cross‑language similarity.

\section{Methodology}

The main hypothesis in this work is that Large Language Models are a good choice to measure internal language structure since they are trained to model languages. Formally, this is typically framed as maximizing the log-likelihood of the observed sequence of tokens. Let $x_1, x_2, \dots, x_T$ represent a sequence of tokens, where $x_t \in \mathcal{V}$ and $\mathcal{V}$ is the vocabulary. The objective is to maximize the likelihood of the sequence under the model's parameters $\theta$:

\[
\mathcal{L}(\theta) = \sum_{t=1}^{T} \log p(x_t | x_1, x_2, \dots, x_{t-1}; \theta),
\]
where \( p(x_t | x_1, x_2, \dots, x_{t-1}; \theta) \) is the conditional probability of the token $x_t$ given the previous tokens, modeled by a neural network or another probabilistic model.


\subsection{Weight Importance Metric}

We begin by revisiting classical pruning approaches such as Optimal Brain Damage \cite{lecun1989optimal}, which motivate the rationale behind our approach.

The typical pruning objective is to minimize the error introduced by approximating the original weight matrix. Consider the following objective function:
\begin{equation}
\label{eq:pruning_objective}
    E = \| \mathbf{W}\mathbf{X} - \mathbf{\hat{W}}\mathbf{X} \|_2^2 \rightarrow \min,
\end{equation}
where \( \mathbf{W} \) is the original weight matrix of a layer, \( \mathbf{\hat{W}} \) is the pruned (sparse) weight matrix, and \( \mathbf{X} \) is the input to that layer.

The variation of the error \( E \) for a weight row \( \mathbf{w} \) can be expressed as:
\[
\delta E = \left( \frac{\partial E}{\partial \mathbf{w}} \right)^T \delta \mathbf{w}
+ \frac{1}{2} \, \delta \mathbf{w}^T \mathbf{H} \, \delta \mathbf{w} + \mathcal{O}(\|\delta \mathbf{w}\|^3),
\]
where \( \mathbf{H} \equiv \frac{\partial^2 E}{\partial \mathbf{w}^2} \) is the Hessian matrix.

At a local minimum of the training error, we have
\[
\frac{\partial E}{\partial \mathbf{w}} \approx 0,
\]
and higher order terms are neglected.

Our goal is to set one of the weights, say \( w_q \), to zero while minimizing the increase in error. This introduces the constraint:
\[
\mathbf{e}_q^T \delta \mathbf{w} + w_q = 0,
\]
where \( \mathbf{e}_q \) is the \( q \)th standard basis vector. Thus, the optimization problem in Equation~\eqref{eq:pruning_objective} can be reformulated as:
\begin{equation}
\label{eq:hessian_st}
\begin{aligned}
\min_{\delta \mathbf{w}} \quad & \frac{1}{2} \, \delta \mathbf{w}^T \mathbf{H} \, \delta \mathbf{w}, \\
\text{s.t.} \quad & \mathbf{e}_q^T \delta \mathbf{w} + w_q = 0.
\end{aligned}
\end{equation}

This constrained problem can be solved using Lagrange multipliers. For the detailed derivation see Appendix \ref{appendix:weight_importance_metric}.

The resulting increase in error is given by:
\begin{equation}
\label{eq:change_in_error}
E_q = \frac{1}{2} \cdot \frac{w_q^2}{\mathbf{e}_q^\top \mathbf{H^{-1}} \mathbf{e}_q}.
\end{equation}

By computing \( E_q \) for every weight \( w_q \), one can prune the weight that causes the smallest increase in error, thereby minimally affecting the layer's output. Intuitively, this means we identify which weights are most critical for the model's performance on a specific language. Weights with high importance scores are those whose removal would substantially degrade the model's ability to predict tokens in that language.

SparseGPT \cite{frantar2023sparsegpt} adopts this idea within an LLM pruning algorithm.
They compute the importance metric \( \mathbf{S}_{ij} \) for a layer as follows \cite{sun2023wanda}:
\begin{equation}
\label{eq:sparseGPT}
    \mathbf{S}_{ij} = \left[ \frac{|\mathbf{W}|^{2}}{\text{diag}\Bigl( ({\bf X}^{T}{\bf X} + \lambda {\bf I})^{-1} \Bigr)} \right]_{ij}.
\end{equation}

As in SparseGPT, we build \(\mathbf{X}\) \emph{per linear sub‑layer} by stacking the
pre‑activation hidden states of a small calibration set into an
\(N\times d_{\text{in}}\) matrix. For a weight matrix \(\mathbf{W}\) the local Hessian
is \(\mathbf{H} = \mathbf{X}^{\top}\mathbf{X}\), and we invert \(\bigl(\mathbf{X}^{\top}\mathbf{X}+\lambda \mathbf{I}\bigr)\) \emph{once
per layer}. Thus, Equation~\eqref{eq:sparseGPT} is simply a matrix-valued, regularised
version of the scalar error-increase criterion in
Equation~\eqref{eq:change_in_error}. 

\citet{shamrai2024language} suggests that the SparseGPT algorithm provides statistically stable results for different LLMs and subsets of a data in language-specific setting. Therefore, in our work, we adopt the algorithm to compute weight importance vectors.

\subsection{Rationale Behind the Approach}

By definition, \( \mathbf{S}_{ij} \) quantifies the importance of weight \( \mathbf{W}_{ij} \) for a given input. In our approach, we estimate the importance of the weights for a specific language by using datasets in that language. Consequently, \( \mathbf{S}_{ij} \) reflects the contribution of each weight to language modeling.

Assuming that the network is well-trained on language modeling, higher \( \mathbf{S} \) scores indicate greater contribution. If two languages yield similar patterns of important weights, it suggests that they are similar in terms of language modeling characteristics.

\subsection{Constructing a Metric Space}

To derive a vector representation from the importance metric, we treat the importance scores as coordinates in a high-dimensional space. Specifically, we define the vector
\[
\mathbf{v} = \bigl( \mathbf{S}_{00}^0, \mathbf{S}_{01}^0, \dots, \mathbf{S}_{ij}^k, \dots, \mathbf{S}_{nm}^l \bigr) \in \mathbb{R}^{N},
\]
where the set \( \{ \mathbf{W}^k \}_{k=0}^{l} \) consists of weight matrices \( \mathbf{W}^k \in \mathbb{R}^{n_k \times m_k} \) for each layer $k$, and \( N \) is the total number of parameters in the chosen LLM. In other words, the vector \( \mathbf{v} \) is obtained by flattening and concatenating all the importance matrices \( \mathbf{S}^k \) corresponding to each layer.

There are two challenges with using the raw importance matrix \( \mathbf{S} \) to form this vector representation:
\begin{enumerate}
    \item The importance scores are not normalized across layers, meaning that they are only meaningful within the context of a single layer.
    \item The resulting vector is high-dimensional, with each dimension represented by a floating-point number (typically 16 bits), leading to large memory requirements.
\end{enumerate}

To mitigate this, we propose a thresholding approach analogous to binary quantization. Specifically, we assign a value of \( 1 \) only to the most important weights by thresholding \( \mathbf{S}_{ij} \) at its median:
\[
\mathbf{\hat{S}}_{ij} = \mathds{1}\bigl( \mathbf{S}_{ij} > \text{median}(\mathbf{S}) \bigr).
\]
This binary representation requires only 1 bit per value, reducing the storage requirement substantially compared to 16-bit floating-point representations.

Let \( X \) denote the set of language vectors (one per language) of length \(N\). We then define a metric space on \( X \) using the Hamming distance (i.e., the XOR operation) as the metric.

For \(\mathbf{x},\mathbf{y}\in X\) the Hamming distance is

\[
d_h(\mathbf{x},\mathbf{y})
\;=\;
\sum_{i=1}^{N} 
\mathds{1}\!\bigl[x_i \neq y_i\bigr],
\]

where \(\mathds{1}[\cdot]\) is the indicator function.

The function \(d_h\) is non‑negative, symmetric, equals \(0\) iff \(\mathbf{x}=\mathbf{y}\), and satisfies the triangle inequality, therefore, \((X,d_h)\) is a metric space.

\begin{algorithm}[ht]
\caption{Torgerson Scaling (Classical MDS)}
\label{alg:classical_mds}
\begin{algorithmic}[1]
\Require Distance matrix $D \in \mathbb{R}^{n \times n}$, $n = |X|$
\Ensure Coordinates $Y \in \mathbb{R}^{n \times d}$ representing points in $d$ dimensions
\State $J \gets I_n - \frac{1}{n}\mathbf{1}_n$ \Comment{Compute centering matrix}
\State $D^2 \gets D \odot D$ \Comment{Element-wise square of $D$}
\State $B \gets -\frac{1}{2} \, J \, D^2 \, J$ \Comment{Compute Gram matrix}
\State $(\lambda, V) \gets \texttt{eigh}(B)$ \Comment{Compute the eigen-decomposition of $B$}
\State $(\lambda, V) \gets \texttt{sort}((\lambda, V))$ \Comment{Sort eigenvalues in descending order and reorder eigenvectors accordingly}
\State $d \gets \#\{ \lambda_i \mid \lambda_i > \epsilon \}$ \Comment{Select dimensions with significant eigenvalues ($\epsilon \approx 10^{-10}$)}
\State $L \gets \mathrm{diag}(\sqrt{\lambda_1}, \sqrt{\lambda_2}, \dots, \sqrt{\lambda_d})$
\State $V_d \gets [v_1, v_2, \dots, v_d]$
\State \Return $Y \gets V_d \, L$
\end{algorithmic}
\end{algorithm}

\subsection{Isometry via Dimensionality Reduction}

Even after quantization, the binary vectors remain high-dimensional due to the large number of model parameters, making distance computations and other latent space applications computationally expensive. To address this, we construct an isometry -- a transformation that preserves distances between points when mapping from one metric space to another.

In our experiments, we employ different LLMs and multiple datasets. We compute the language-by-language distance matrix for each model and dataset, and then average them to obtain a robust distance measure:

\begin{table*}[t]
\centering
\begin{tabular}{lcccc}
\toprule
Dataset    & \# Languages in Dataset & \# Languages Used in Work \\
\midrule
Wikipedia  & 323                    & 106 \\
CulturaX   & 167                    & 102 \\
fineweb-2  & 2051                  & 103  \\
\bottomrule
\end{tabular}
\caption{Comparison of datasets: Wikipedia, CulturaX, and fineweb-2. The table reports the total number of languages in each dataset and the number of languages used in this work.}
\label{tab:dataset_comparison}
\end{table*}

\begin{align*}
    \mathbf{D}_{lk} &\in \mathbb{R}^{|X| \times |X|}, \\ \mathbf{D}_{lk} &= \{ d_{h}(\mathbf{v}_i, \mathbf{v}_j) : \mathbf{v}_i, \mathbf{v}_j \in X \}, \\[10pt]
    \mathbf{\hat{D}} &= \mathbb{E}_{l\sim p_{\mathrm{LLM}}}\mathbb{E}_{k\sim p_{\mathrm{data}}}[\mathbf{D}_{lk}] \\
    &\approx \frac{1}{nm}\sum_{l=0}^{n}\sum_{k=0}^{m} \mathbf{D}_{lk}, \\
\end{align*}
where \( \mathbf{D}_{lk} \) is the distance matrix computed for the \( l \)th LLM and the \( k \)th dataset, \( n \) is the number of LLMs, \( m \) is the number of datasets, and $|X|$ is the number of languages.

This averaging process reduces noise and ensures that the final distances are not overly dependent on any particular dataset or model.

We then construct an isometry
\[
f: X \rightarrow Y,
\]
where \( Y \) is a metric space endowed with the Euclidean metric \( d_{e}(x, y) = \| x - y \|_2 \). 

To build \( f \), we apply Torgerson scaling (classical multidimensional scaling) \cite{borg2007modern}. The result is a set of points \( Y \in \mathbb{R}^{|X| \times d} \), where \( d \) is the minimum number of dimensions required to preserve the distances in \( \mathbf{\hat{D}} \) (see Algorithm~\ref{alg:classical_mds}). Notably, \( d \) is much smaller than the original dimensionality \( N \) of the language vectors and satisfies \( d \leq |X| \).

\vspace{0.5cm}

Therefore, our method leverages LLMs weights to construct a language vector representation and embed it in a metric space which could be used for analysis of languages similarities.

\section{Results}

To analyze the metric space of languages, we vary clustering algorithms along with dimensionality reduction ones. In particular, for clustering HDBSCAN \cite{Campello2013Density}, $k$-means \cite{Lloyd1982Least}, and predefined linguistic families with its subfamilies are used to highlight the correspondence between the derived metric space and established linguistic classifications. Throughout this paper we adhere to the language classification provided in \citet{glottolog51}.

For two‑dimensional visualizations, we reduce the dimensionality of the language vectors using t-SNE \cite{vandermaaten08a}, UMAP \cite{mcinnes2018umap}, and minimum spanning trees \cite{pettie2002optimal}. Although all methods yield valuable insights, we include in the main text only the minimum spanning trees (MST) visualizations colored by language families and subfamilies, as they most clearly represent the inter-language relationships. 
Additional figures are provided in Appendix \ref{appendix:plots} and also available via our open‑source tool\footnote{\url{https://huggingface.co/spaces/mshamrai/language-metric-analysis}}.

\subsection{Datasets and Models}

In our experiments, we employ three LLMs and three datasets. The models used are Mistral 7B \cite{jiang2023mistral}, Gemma 3 4B \cite{team2025gemma}, and Llama 3.2 1B \cite{grattafiori2024llama}. All models are multilingual and have been trained on more than 100 languages. Notably, although Llama officially supports only 8 languages, our results indicate that it still produces useful representations for our purposes. As datasets, we selected those with a high number of languages: Wikipedia \cite{wikidump}, CulturaX \cite{nguyen-etal-2024-culturax}, and fineweb-2 \cite{penedo2024fineweb-2}.

We start with a target inventory of 106 languages and attempted to apply
the same list across all corpora.  Wikipedia contains material for every
language in this set, but CulturaX omits Chinese (Traditional), Min Nan Chinese, Scots, and
Crimean Tatar, whereas fineweb-2 lacks Chinese (Traditional), English\footnote{For the English subset, we use the \textit{fineweb} dataset
(\url{https://huggingface.co/datasets/HuggingFaceFW/fineweb})},
Serbo-Croatian, and Tagalog.  
Table~\ref{tab:dataset_comparison} lists the total number of languages
present in each dataset alongside the subset that could be retained from
our 106-language list. For the full list of languages see Appendix \ref{appendix:languages}. 


To compute the language vectors, we proceed as follows:

\begin{enumerate}
\item \textbf{Calibration data.}
For every language in each corpus (Wikipedia, CulturaX, fineweb) we sample $2^{19}=524{,}288$ tokens.

\item \textbf{Weight‑importance vectors.}
For each language–corpus pair and for each LLM (Mistral 7B, Gemma 3 4B, Llama 3.2 1B) we compute a binary weight importance vector whose length matches the model’s parameter count, yielding $3(106+102+103)=933$ vectors.

\item \textbf{Distance matrices.}
Hamming distances between language vectors produce nine language–by–language matrices (one per model–corpus combination).

\item \textbf{Aggregation.}
These nine matrices are averaged element‑wise over the observed entries to form a single average distance matrix.

\item \textbf{Embedding.}
Classical MDS on the average matrix embeds the languages space in $\mathbb{R}^{104}$, where Euclidean distance defines the final language metric.
\end{enumerate}

\subsection{Evaluation of $k$‑means Clustering Against Two Linguistic Categorization}

After we embed the $|X|=106$ language vectors into $\mathbb{R}^{104}$ via classical MDS we evaluate the language embeddings using $k$‑means.
The resulting partition is compared with two reference label sets:
(i) \textit{high‑level families} (18 macro‑families) and  
(ii) \textit{primary branches} (35 sub‑families).
The number of clusters in $k$‑means is equal to the number of labels in the reference sets. 

We compute the following metrics:
\begin{itemize}
    \item \textbf{Silhouette score} \cite{rousseeuw1987silhouettes}: the mean difference between a point’s average distance to its own cluster and to the nearest neighboring cluster. Values range from $-1$ (poor separation) to $+1$ (well‑separated, compact clusters).
    \item \textbf{Adjusted Rand Index (ARI)} \cite{hubert1985comparing}: agreement between two partitions, corrected for chance. $1$ indicates perfect alignment, $0$ indicates random overlap.
    \item \textbf{Cluster purity} \cite{schutze2008introduction}: the fraction of data points that share the majority label within their cluster. Values in $[0,1]$.
\end{itemize}

\begin{table}[ht]
    \centering
    \begin{tabular}{lcccc}
        \toprule
        Reference          & Sil. & ARI    & Purity \\
        \midrule
        Macro families       & 0.047 & 0.116 & 0.755 \\
        Primary branches     & 0.056 & 0.434 & 0.811 \\
        \bottomrule
    \end{tabular}
    \caption{Clustering metrics for the $k$‑means solution against two standard language classification. "Sil" is the internal silhouette score.}
    \label{tab:clustering_metrics}
\end{table}

Table~\ref{tab:clustering_metrics} shows that switching from broad families to primary branches raises the ARI from~0.116 to~0.434 and the purity from~0.755 to~0.811.  
Therefore, the metric space captures finer‑grained language groups and can estimate similarity at a micro level.  
However, the internal silhouette remains low (about~0.05), meaning many languages lie almost as close to other clusters as to their own. 

\subsection{Language Trees}

\begin{figure*}[htp]
    \centering
    \includegraphics[width=\linewidth]{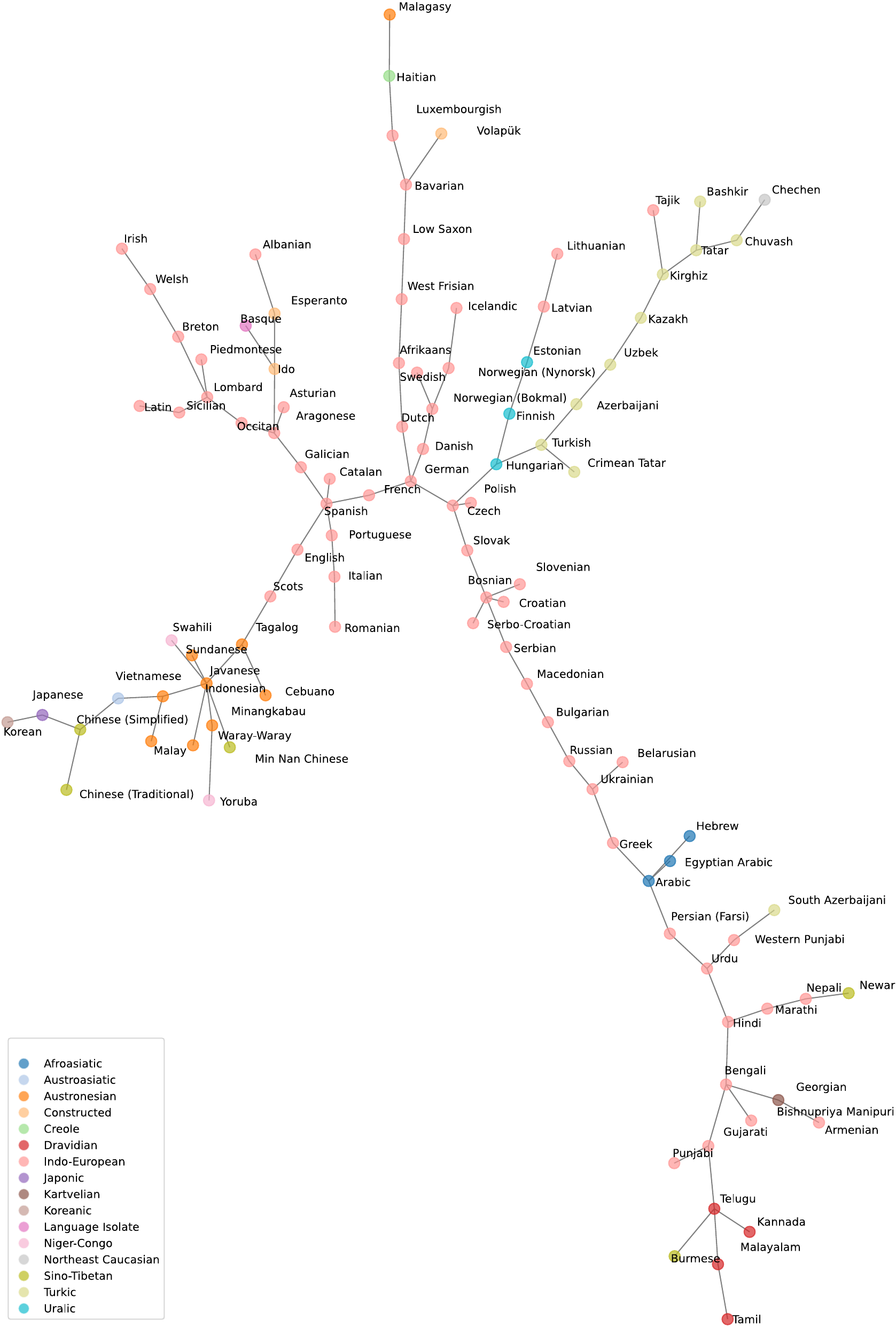}
    \caption{Minimum spanning tree for all languages. Colors represent language families.}
    \label{fig:all_families}
\end{figure*}

\begin{figure*}[ht]
    \centering
    \includegraphics[width=\linewidth]{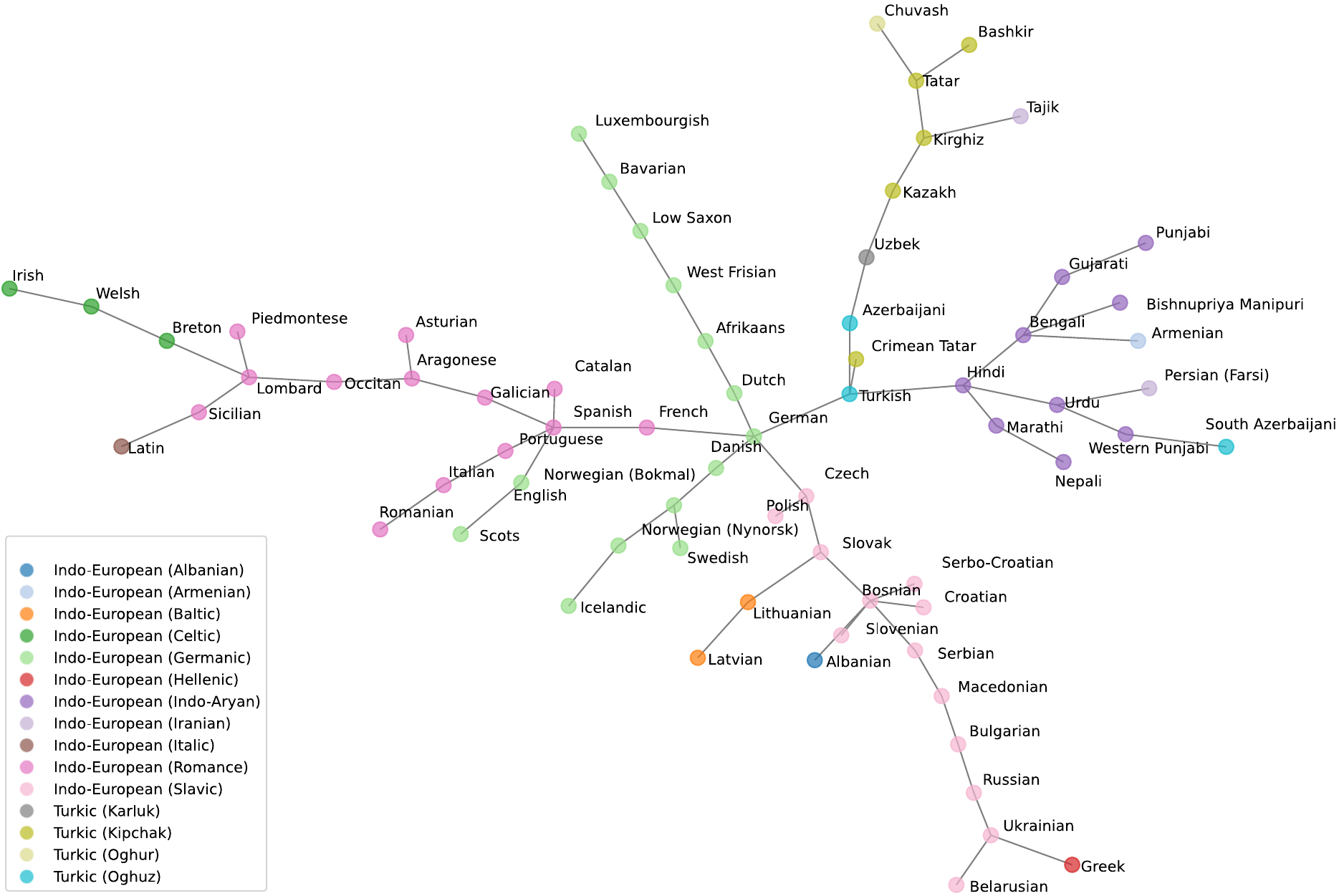}
    \caption{Minimum spanning tree for languages from the Indo-European and Turkic families. Colors represent language primary branches.}
    \label{fig:ie_turkic_families}
\end{figure*}

A minimum spanning tree (MST) connects all data points in the dataset with the smallest possible total edge weight, where the edge weight corresponds to the distance between language vectors. We employ the Kamada-Kawai layout, a force-directed algorithm where edge lengths are proportional to the distances \cite{kamada1989algorithm}. This layout effectively visualizes the structure and connectivity within the MST, revealing not only the clusters of closely related languages but also links between different language families.

Figure \ref{fig:all_families} shows the MST for all languages used in our work. The visualization highlights well-established clusters corresponding to known language families as well as some unexpected connections. For example, Tajik (an Indo-European language) appears linked to a cluster of Turkic languages, which can likely be explained by geographical proximity. Similarly, the branch containing Latvian and Lithuanian is connected to a cluster of Uralic languages, possibly due to regional contact with Finnish and Estonian. A less obvious connection is observed between Turkish and Hungarian, which might be attributed to historical interactions. Additionally, Vietnamese is found to be close to Chinese, despite Vietnamese using the Latin alphabet and Chinese employing logographic characters, indicating that our method captures internal language characteristics beyond mere orthographic features.

Figure \ref{fig:ie_turkic_families} focuses on Indo-European and Turkic languages, with coloring based on their primary branches. This figure clearly illustrates that Crimean Tatar, although belonging to the Kipchak branch, is closely connected to Turkish, an Oghuz language. The MST also links English, a Germanic language, directly to Spanish, a Romance language, likely reflecting their close geographic and sociolinguistic contact in the Americas. 

One intriguing observation is that Ukrainian does not exhibit a direct connection with Polish in the MST, which is unexpected. However, further analysis reveals that Polish consistently ranks among the top five closest languages to Ukrainian across all models and datasets, coming in third after averaging the distances.

\vspace{0.5cm}

In summary, the minimum spanning trees reveal logical relationships among languages and their families. In addition, the presence of uncommon connections suggests potential historical contacts or convergent evolution. We leave further investigation of areal influences or language borrowing phenomena to professionals.

\section{Conclusion}

In this work, we introduced a novel framework for constructing a metric space that quantifies language similarity by leveraging the internal weight activations of Large Language Models. 

Our approach, based on computing binary vectors from weight importance metrics and reducing their dimensionality via isometric mappings, captures linguistic features, and the resulting metric space not only aligns with established linguistic families but also reveals intriguing inter-language connections.

Overall, this study lays the groundwork for a data-driven paradigm in language similarity analysis with significant implications for theoretical linguistics.

\section*{Limitations}

While our approach offers a novel perspective on constructing a metric space for languages using LLM weight activations, several limitations remain:

\begin{enumerate}
    \item \textbf{Computational Expense:} Computing the binary vectors is time-consuming. For example, on the Mistral 7B model, generating one binary vector requires approximately 20 minutes on an NVIDIA RTX 3090 GPU.
    \item \textbf{Scalability to Larger Models:} We have not yet evaluated the method on LLMs with a significantly higher number of parameters due to resource constraints. It is possible that larger models might yield more accurate or robust representations.
    \item \textbf{Remaining Bias from Source Models:} Averaging distances across three LLMs does not eliminate their shared weaknesses. In particular, the metric space can still reflect poor performance on low resource languages, which may introduce inconsistencies with known language family relationships.
\end{enumerate}

Additionally, we were unable to mathematically or empirically validate that the derived distance metric can serve as an effective guideline for fine-tuning and transfer learning of LLMs. Although the underlying hypothesis suggests that linguistic similarity may enhance the language modeling capabilities through transfer learning between related languages, our preliminary experiments -- where we fine-tuned an LLM on similar languages using various configurations -- did not yield statistically significant improvements. This indicates that a more sophisticated approach may be required, and we leave this investigation for future work. For more details see Appendix \ref{appendix:tl}.

Another promising direction for future research is to identify which specific weights or layers contribute most to the observed similarities and dissimilarities. It is likely that only a subset of layers significantly influences the metric. By pinpointing these layers, we may reduce computational complexity and accelerate the metric computation without compromising accuracy.

\section*{Acknowledgment}
We express our gratitude to the Armed Forces of Ukraine for their protection, which has made this research possible.

\bibliography{ranlp2025}

\appendix

\section{Derivation of weight importance metric}
\label{appendix:weight_importance_metric}

\begin{equation*}
\begin{aligned}
\min_{\delta \mathbf{w}} \quad & \frac{1}{2} \delta \mathbf{w}^T \mathbf{H} \, \delta \mathbf{w}, \\
\text{s.t.} \quad & \mathbf{e}_q^T \delta \mathbf{w} + w_q = 0.
\end{aligned}
\end{equation*}

The problem could be solved using Lagrange multiplier. 
We begin with the Lagrangian:
\begin{equation*}
\mathcal{L} = \frac{1}{2} \delta \mathbf{w}^\top \mathbf{H} \delta \mathbf{w} 
+ \lambda \left( \mathbf{e}_q^\top \delta \mathbf{w} + w_q \right).
\end{equation*}

Taking the derivative with respect to \( \delta \mathbf{w} \) and setting it to zero:
\begin{align*}
\nabla_{\delta \mathbf{w}} \mathcal{L} &= \mathbf{H} \delta \mathbf{w} + \lambda \mathbf{e}_q = 0, \\
\Rightarrow \quad \delta \mathbf{w} &= -\mathbf{H}^{-1} \mathbf{e}_q \lambda.
\end{align*}

Substituting into the constraint:
\begin{equation*}
\mathbf{e}_q^\top (-\mathbf{H}^{-1} \mathbf{e}_q \lambda) + w_q = 0,
\end{equation*}
we get:
\begin{equation*}
\lambda = \frac{w_q}{\mathbf{e}_q^\top \mathbf{H}^{-1} \mathbf{e}_q}.
\end{equation*}

Thus, the change in weights:
\begin{equation*}
\delta \mathbf{w} = -\mathbf{H}^{-1} \mathbf{e}_q \cdot \frac{w_q}{\mathbf{e}_q^\top \mathbf{H}^{-1} \mathbf{e}_q}.
\end{equation*}

Notice that:

\begin{align*}
\mathbf{H} \delta \mathbf{w} &= \mathbf{H}
\left(
- \mathbf{H}^{-1} \mathbf{e}_q \frac{w_q}{\mathbf{e}_q^\top \mathbf{H}^{-1} \mathbf{e}_q}
\right) \\
&= - \mathbf{e}_q \frac{w_q}{\mathbf{e}_q^\top \mathbf{H}^{-1} \mathbf{e}_q},
\end{align*}
and
\begin{align*}
\delta \mathbf{w}^\top &= \left( 
- \mathbf{H}^{-1} \mathbf{e}_q \frac{w_q}{\mathbf{e}_q^\top \mathbf{H}^{-1} \mathbf{e}_q}
\right)^\top \\
&= - \frac{w_q}{\mathbf{e}_q^\top \mathbf{H}^{-1} \mathbf{e}_q} \mathbf{e}_q^\top \mathbf{H}^{-1}
\end{align*}

Now compute the increase in error:
\begin{align*}
E_q &= \frac{1}{2} \delta \mathbf{w}^\top \mathbf{H} \delta \mathbf{w} \\
&= \frac{1}{2} \frac{w_q}{\mathbf{e}_q^\top \mathbf{H}^{-1} \mathbf{e}_q} \mathbf{e}_q^\top \mathbf{H}^{-1} \mathbf{e}_q \frac{w_q}{\mathbf{e}_q^\top \mathbf{H}^{-1} \mathbf{e}_q} \\
&= \frac{1}{2} \cdot \frac{w_q^2}{\mathbf{e}_q^\top \mathbf{H}^{-1} \mathbf{e}_q}
\end{align*}

\section{Full list of languages used}
\label{appendix:languages}

Afrikaans, Albanian, Arabic, Aragonese, Armenian, Asturian, Azerbaijani, Bashkir, Basque, Bavarian, Belarusian, Bengali, Bishnupriya Manipuri, Bosnian, Breton, Bulgarian, Burmese, Catalan, Cebuano, Chechen, Chinese (Simplified), Chinese (Traditional), Chuvash, Crimean Tatar, Croatian, Czech, Danish, Dutch, Egyptian Arabic, English, Esperanto, Estonian, Finnish, French, Galician, Georgian, German, Greek, Gujarati, Haitian, Hebrew, Hindi, Hungarian, Icelandic, Ido, Indonesian, Irish, Italian, Japanese, Javanese, Kannada, Kazakh, Kirghiz, Korean, Latin, Latvian, Lithuanian, Lombard, Low Saxon, Luxembourgish, Macedonian, Malagasy, Malay, Malayalam, Marathi, Min Nan Chinese, Minangkabau, Nepali, Newar, Norwegian (Bokmal), Norwegian (Nynorsk), Occitan, Persian (Farsi), Piedmontese, Polish, Portuguese, Punjabi, Romanian, Russian, Scots, Serbian, Serbo-Croatian, Sicilian, Slovak, Slovenian, South Azerbaijani, Spanish, Sundanese, Swahili, Swedish, Tagalog, Tajik, Tamil, Tatar, Telugu, Turkish, Ukrainian, Urdu, Uzbek, Vietnamese, Volapük, Waray-Waray, Welsh, West Frisian, Western Punjabi, Yoruba.

Wikipedia includes all these languages. CulturaX lacks Chinese (Traditional), Min Nan Chinese, Scots, and Crimean Tatar. fineweb-2 does not include Chinese (Traditional), English, Serbo-Croatian, or Tagalog. For the English subset in fineweb-2, we use the fineweb dataset\footnote{\url{https://huggingface.co/datasets/HuggingFaceFW/fineweb}}.

\section{Transfer‑Learning Experiments}
\label{appendix:tl}

We investigated whether adding data from a \emph{similar} language can improve a low‑resource target, where similarity is measured by the language‑distance metric introduced in this paper.
All experiments fine‑tune Llama 3.2 1B and evaluate exclusively on a held‑out set in the target language.

We perform our experiments using the following strategies:

\begin{enumerate}
\item \textbf{Mixed (size‑matched).}  
      An equal amount of auxiliary‑language text is concatenated to the low‑resource corpus; the joint data are shuffled and used for fine‑tuning.

\item \textbf{Mixed (loss‑weighted).}  
      The same joint corpus is used, but the loss is re‑weighted: e.g. $0.8$ for target‑language tokens and $0.2$ for auxiliary‑language tokens.

\item \textbf{Sequential.}  
      Fine‑tune first on the auxiliary language, then continue training on the low‑resource corpus.
\end{enumerate}

\begin{figure}[t]
  \centering
  \includegraphics[width=\linewidth]{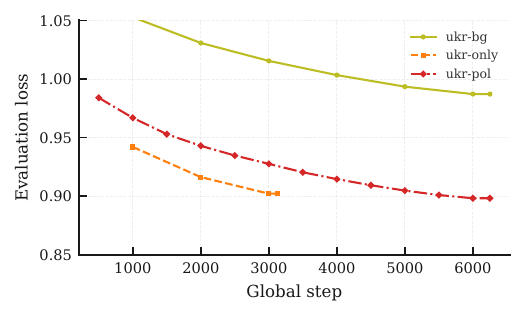}
  \caption{Evaluation loss on Ukrainian for three weighted‑loss runs: 
           \textit{ukr‑only} (baseline), \textit{ukr‑bg} 
           (Ukrainian + Bulgarian), and \textit{ukr‑pol} (Ukrainian + Polish).  
           Two‑language datasets are twice as large, hence the longer training schedule.}
  \label{fig:eval_loss}
\end{figure}

Figure \ref{fig:eval_loss} shows that augmenting Ukrainian with the metrically close Bulgarian does not improve evaluation loss, and Polish yields only a minor reduction.  

A similar pattern emerges for sequential fine‑tuning on Turkish followed by Crimean Tatar: perplexity drops from 5.48 (Crimean Tatar only) to 5.36, an insignificant change.

Across all settings, none of the three transfer regimes produced a consistent, significant gain over single‑language fine‑tuning. Future work should revisit these transfer strategies with substantially larger models and much larger datasets, where the benefits of distance‑based language pairing may emerge more clearly.

\section{Additional Figures}
\label{appendix:plots}

\begin{figure*}[t]
    \centering
    \includegraphics[width=\linewidth]{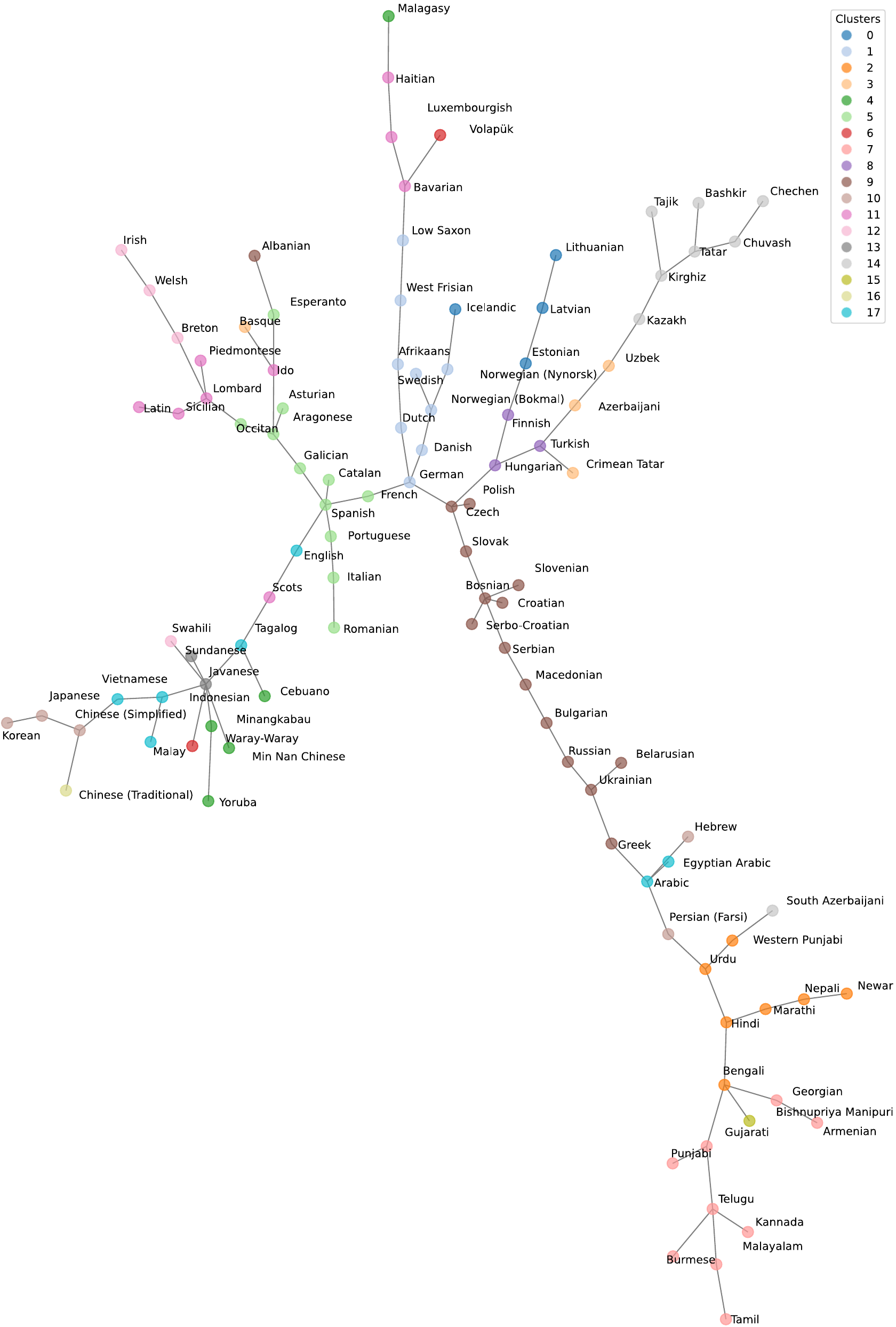}
    \caption{MST of all languages. Colours show $k$‑means clusters with $k=18$ (one cluster for each language family).}
    \label{fig:mst_kmeans_18}
\end{figure*}

Figure~\ref{fig:mst_kmeans_18} displays the MST coloured by $k$‑means clusters.  
We set $k=18$ -- one cluster for each category plotted in
Figure~\ref{fig:all_families} (15 natural families plus 3
constructed languages) -- so that the cluster colours can be compared
directly with the family colours. Most clusters coincide with their
expected families, but not all. Notably, Turkish is grouped with
Hungarian and Finnish rather than with the other Turkic languages.

\begin{figure*}[t]
    \centering
    \includegraphics[width=\linewidth]{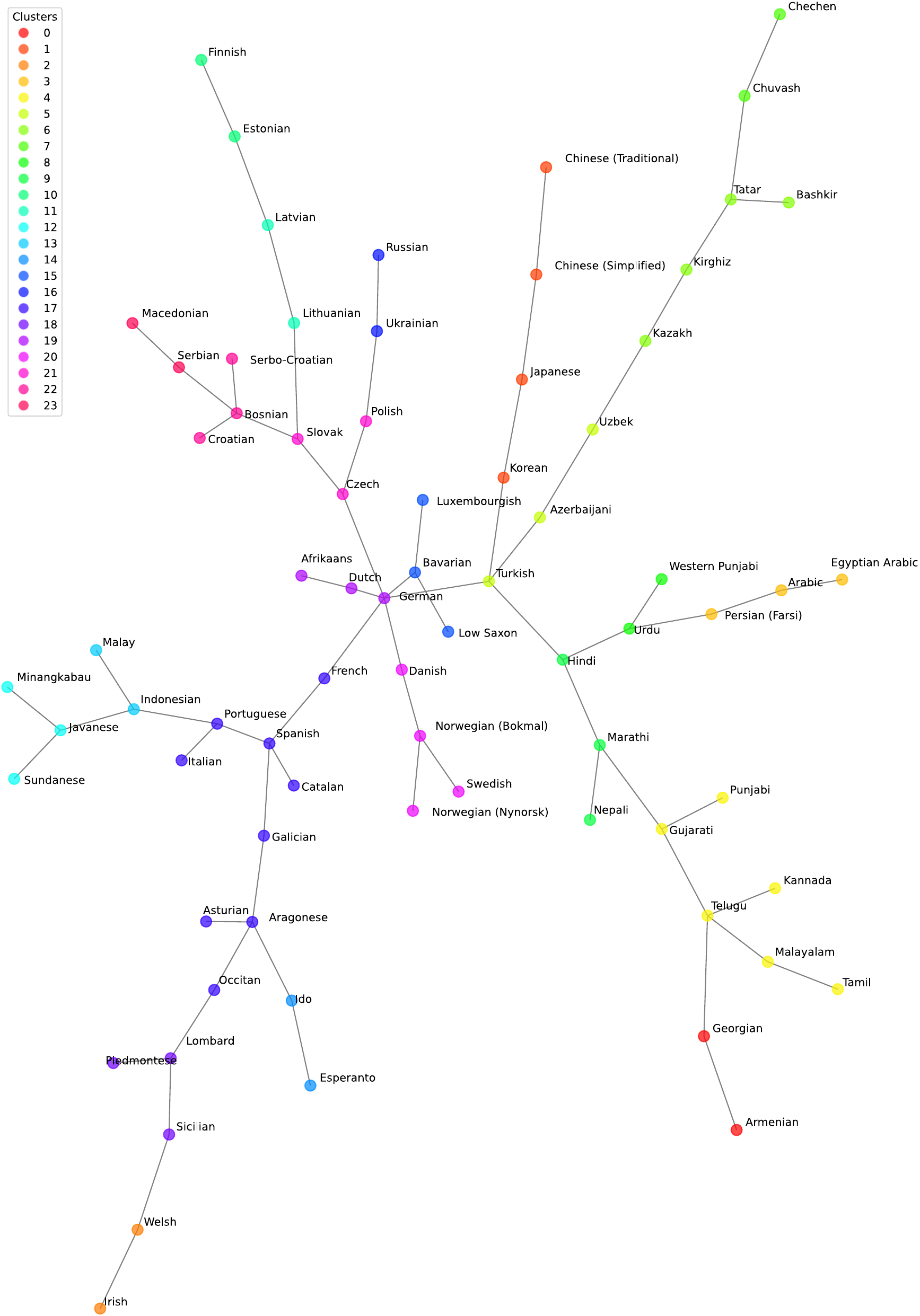}
    \caption{MST of all languages. Colours show HDBSCAN clusters (minimum cluster size = 2). Points marked as outliers by the algorithm are left out.}
    \label{fig:mst_hdbscan_2}
\end{figure*}

Figure~\ref{fig:mst_hdbscan_2} uses HDBSCAN with a minimum cluster size of two.  
This gives 24 clusters.  
Crimean Tatar is treated as outlier, while Ukrainian now connects directly to Polish.

\begin{figure*}[t]
    \centering
    \includegraphics[width=\linewidth]{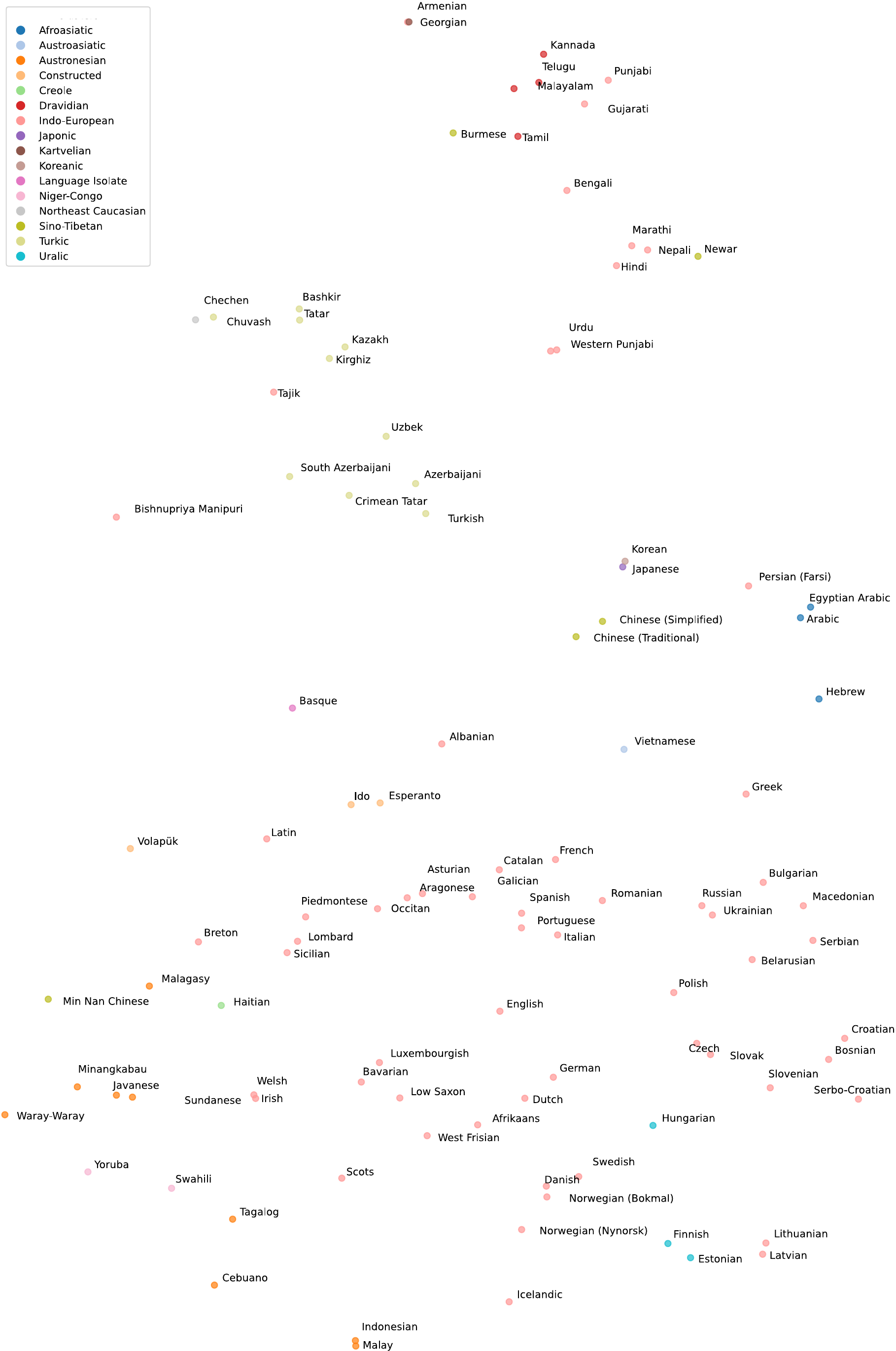}
    \caption{t‑SNE plot of all languages. Colours show language families.}
    \label{fig:tsne_families}
\end{figure*}

\begin{figure*}[t]
    \centering
    \includegraphics[width=\linewidth]{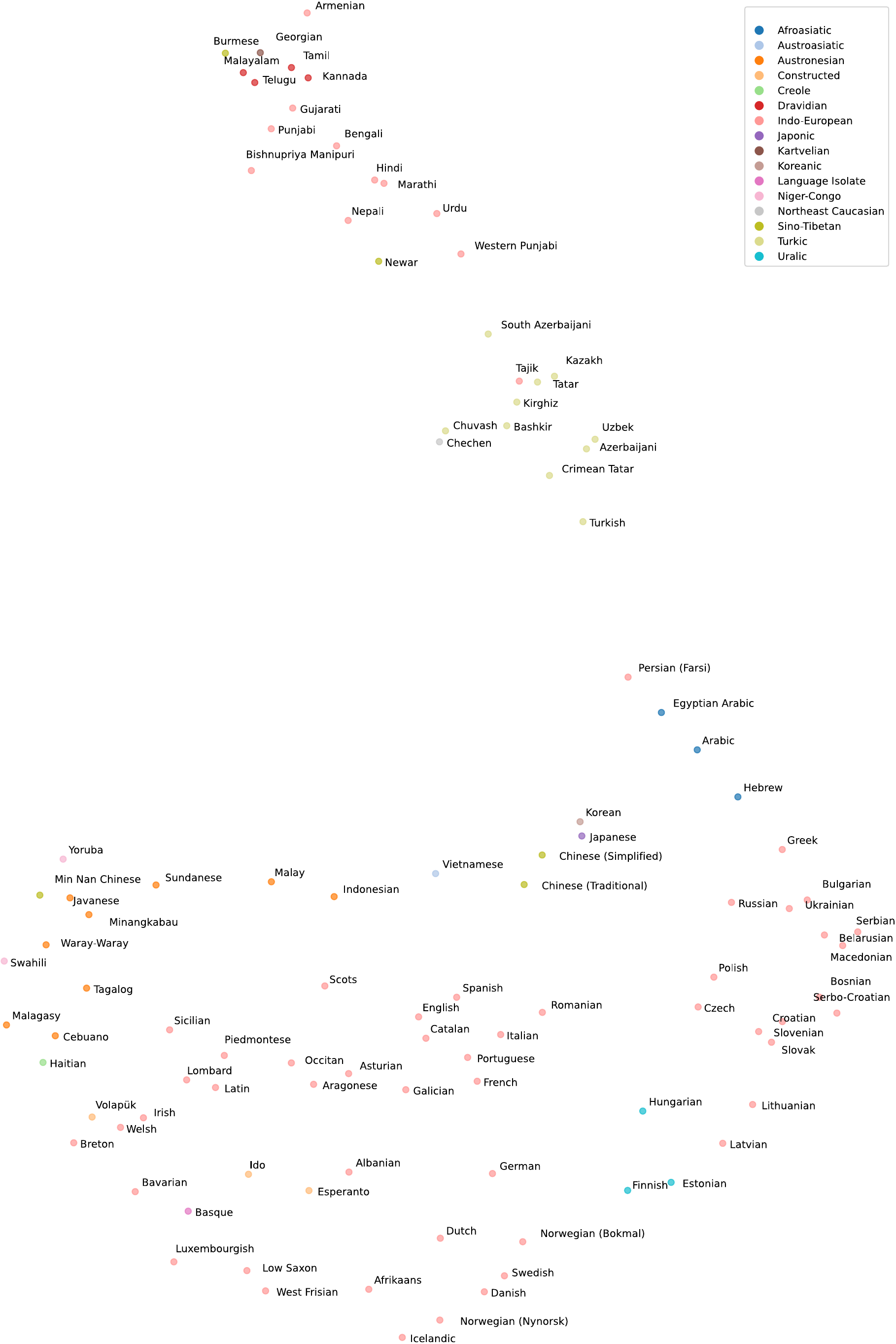}
    \caption{UMAP plot of all languages. Colours show language families.}
    \label{fig:umap_families}
\end{figure*}

Figures~\ref{fig:tsne_families} and~\ref{fig:umap_families} give two other views of the same data using t‑SNE and UMAP.  
Like the MST, they highlight clear family groups.

\begin{figure*}[t]
    \centering
    \includegraphics[width=\linewidth]{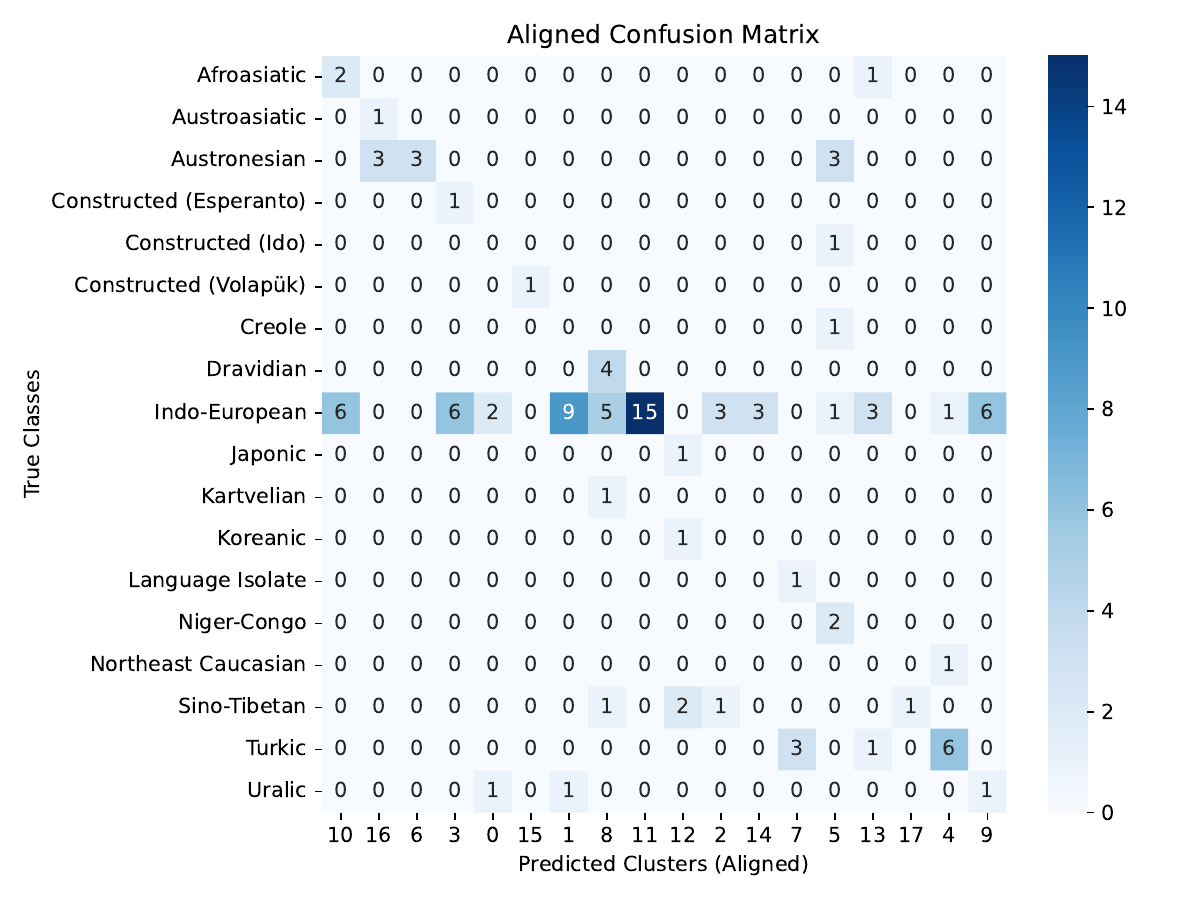}
    \caption{Adjusted confusion matrix between clusters obtained by $k$-means and macro families of languages. Number of clusters equal to 18.}
    \label{fig:high_level_families}
\end{figure*}

\begin{figure*}[t]
    \centering
    \includegraphics[width=\linewidth]{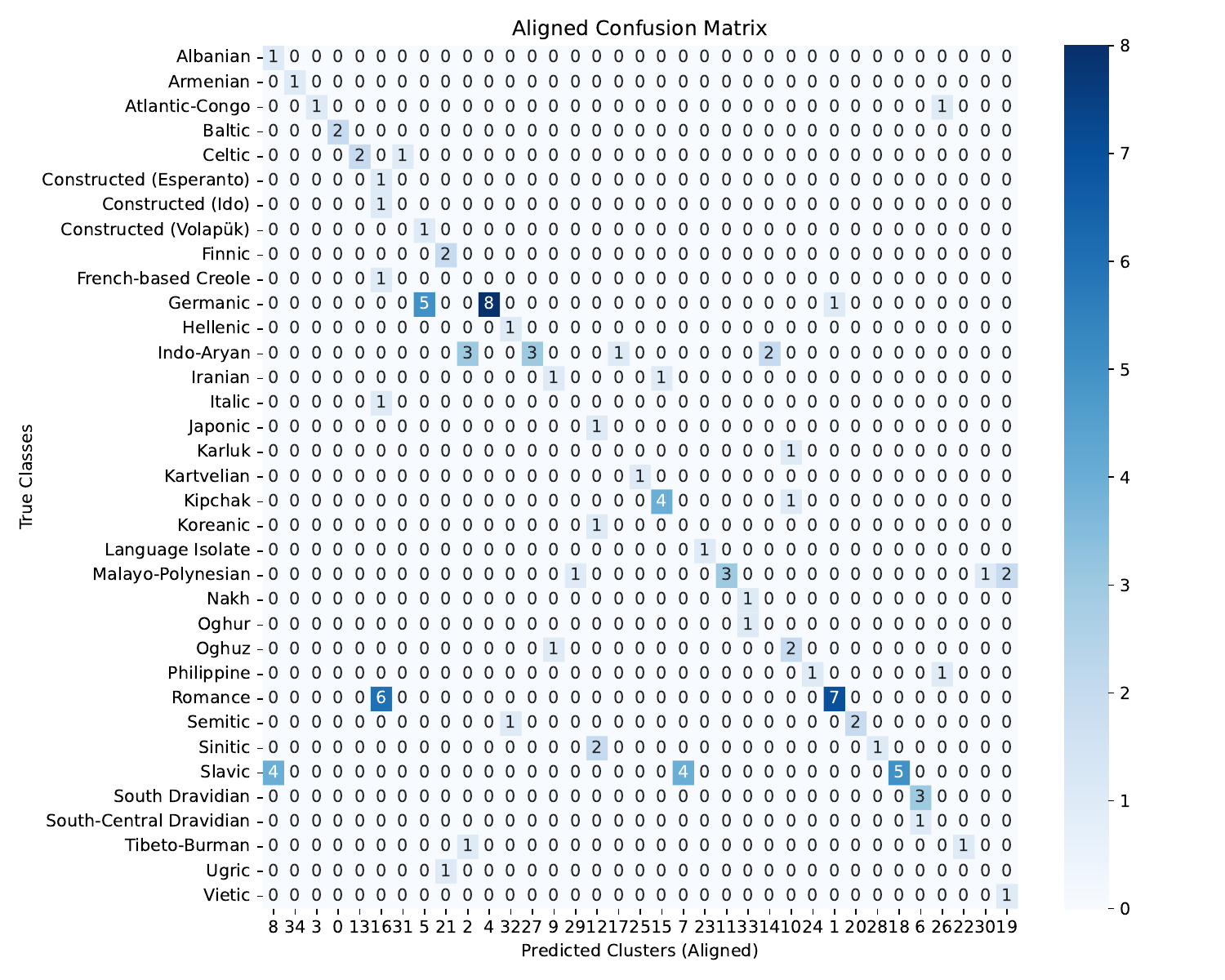}
    \caption{Adjusted confusion matrix between clusters obtained by $k$-means and primary branches of language families. Number of clusters equal to 35.}
    \label{fig:primary_families_branches}
\end{figure*}

Figure~\ref{fig:high_level_families} shows the confusion matrix between $k$‑means clusters and high‑level language families.  
The clusters are first matched to families with the Hungarian algorithm for clearer alignment.  
Figure~\ref{fig:primary_families_branches} presents the same matrix, but for the finer primary branches of each family.

\end{document}